\documentclass[graybox]{svmult}

\usepackage{type1cm}        
\usepackage{makeidx}        
\usepackage{graphicx}       
                            
\usepackage{multicol}       
\usepackage[bottom]{footmisc}

\usepackage{newtxtext}       
\usepackage[varvw]{newtxmath}

\usepackage[T1]{fontenc}

\makeindex

\begin{document}

\title*{Signatures to help interpretability of anomalies}
\author{Emmanuel Gangler\orcidID{0000-0001-6728-1423}, \\
Emille E. O. Ishida \orcidID{0000-0002-0406-076X}, \\
Matwey V. Kornilov \orcidID{0000-0002-5193-9806}, \\
Vladimir Korolev \orcidID{0009-0008-7691-6142}, \\
Anastasia Lavrukhina
\orcidID{0009-0002-0835-7998}, 
\\
Konstantin Malanchev \orcidID{0000-0001-7179-7406}, \\
Maria V. Pruzhinskaya \orcidID{0000-0001-7178-0823}, \\
Etienne Russeil \orcidID{0000-0001-9923-2407}, \\
Timofey Semenikhin \orcidID{0009-0003-2133-6144}, \\
Sreevarsha Sreejith \orcidID{0000-0002-6423-1348} \\
and Alina A. Volnova \orcidID{0000-0003-3554-1037}, \\
}
\authorrunning{Emmanuel Gangler and the SNAD team}

\institute{Emmanuel Gangler (corresponding author), Emille E. O. Ishida, Maria V. Pruzhinskaya and Etienne Russeil \at Université Clermont Auvergne, CNRS, LPCA, 63000 Clermont-Ferrand, France, \email{emmanuel.gangler@clermont.in2p3.fr}
\and Matwey V. Kornilov \at National Research University Higher School of Economics, 21/4 Staraya
Basmannaya Ulitsa, Moscow 105066, Russia
\and Etienne Russeil \at The Oskar Klein Centre, Department of Astronomy, Stockholm University, AlbaNova, SE-10691 Stockholm, Sweden
\and Timofey Semenikhin \at Faculty of Physics, Lomonosov Moscow State University, Leninsky Gori 1, Moscow 119234, Russia
\and Konstantin Malanchev \at McWilliams Center for Cosmology and Astrophysics, Department of Physics,
Carnegie Mellon University, Pittsburgh, PA 15213, USA
\and Timofey Semenikhin, Matwey V. Kornilov, Anastasia Lavrukhina and Maria V. Pruzhinskaya \at Sternberg Astronomical Institute, Lomonosov Moscow State University,
Universitetsky 13, Moscow 119234, Russia
\and Sreevarsha Sreejith \at Physics department, University of Surrey, Stag Hill, Guildford GU2 7XH, UK
\and Alina A. Volnova \at Space Research Institute of the Russian Academy of Sciences, Profsoyuznaya
84/32, Moscow 117997, Russia
}

\maketitle

\abstract{Machine learning is often viewed as a black box when it comes to understanding its output, be it a decision or a score. Automatic anomaly detection is no exception to this rule, and quite often the astronomer is left to independently analyze the data in order to understand why a given event is tagged as an anomaly. We introduce here idea of anomaly signature, whose aim is to help the interpretability of anomalies by highlighting which features contributed to the decision.} 

\section{Challenges for anomaly detection}

By essence, an anomaly is a pattern that does not conform to expected normal behavior, and is suspected to be generated by a different mechanism.
Understanding whether this different mechanism is leading to novel discoveries or is due to already known processes is the role of the domain expert. 
Finding potential anomalies on the other hand is the realm of Machine Learning techniques, due to the huge data volume of current and upcoming surveys.

Several efficient machine learning techniques for anomaly detection have been proposed. A popular option is the use of 
tree-based methods, such as Isolation Forests (IF), which offers low computational complexity while remaining efficient\cite{liu2008isolation}. Interpreting the results of this model relies in mapping the input features importance to the output score. While some model-agnostic methods are proposed, like Shapely Additive Explanations (SHAP) \cite{lundberg2017unified}, they lack the power of addressing the specificity of the model to produce refined explanations. On the other hand, specific methods for IF have been proposed using the depth within the trees as a metric for feature importance \cite{carletti2020interpretable}. While keeping this general idea, 
we propose here a simpler quantity, called Signature, that has a direct link with the IF score.

In the following, we will present first the dataset used for benchmarking our anomaly characterization metric, the novel idea behind signatures and how to use them in order to perform various tasks, such as visualizing important features or discovering new kind of anomalies.

\section{Dataset}

Finding suitable datasets to explore anomalies in the context of astrophysics is challenging as published datasets are often already curated from data artifacts, or lack a comprehensive list of labeled anomalies. 
In this work, we will use the Nearby Supernova Factory (SNfactory) data set.

The SNfactory has made spectrophotometric observations of Type Ia supernovae taking data with the SuperNova Integral Field Spectrograph mounted on the University of Hawaii 2.2~m telescope at Maunakea \cite{2002SPIE.4836...61A,2004SPIE.5249..146L}. In this work, we rely on their public release of 2485 spectra, corresponding to 171 individual objects, the so-called SNEMO companion data set \cite{2020RNAAS...4...63A}. These spectra have been deredshifted to a common restframe at $z=0$, corrected for Milky Way dust extinction, and normalized to an arbitrary magnitude. They cover a wavelength range from 3305.0~\AA\ to 8586.0 \AA\ in 289 spectral bins of equal resolution. With each spectrum is provided also the spectrum of statistical uncertainties, thus porting to 578 the dimension of the available data for each spectrum. We rescaled these spectra so that the integral of the measured flux is 1. No specific featurization has been used, that is, all 578 original bits of information have been used directly as an input to anomaly detection algorithms.

This dataset has some interesting properties for anomaly detection. 
First, Type Ia supernova spectra undergo a significant variability during the life time of the object, leading to complex patterns of what may be labeled as nominal events. Then, even if they have been curated, there are some remaining data reduction artifacts that may be labeled as anomalies.
In addition, spectra are affected by noise which 
will mask potentially interesting anomalies. 

\section{Isolation Forest and Signature definition}

Isolation forest is a well established algorithm for outlier detection. One may see it as an ensemble method built on a variant of extremely randomized trees, where at each node both the attribute and the cut point are selected randomly \cite{liu2008isolation}. Each isolation tree realizes a partition of the original data set according to a training subsample chosen at random from the original training sample. Let $\{x_{fi}\}$ be the original data set of $i\le N$ elements of dimensionality $f\le D$. Each tree $t$ (with $t\le T$) is built out of a subset $S_t=\{x_{fj}\}_t$ of $n$ elements, which are sampled randomly from the original dataset. 
The tree is build by recursively splitting cells $Z_d^k$ 
of $\mathbb{R}^D$ where the root cell is $Z_0^0=\mathbb{R}^D$, 
$0\le d\le d_{\max}$ represents the depth within the tree and $0\le k < 2^d$ is the index of the cell given the depth $d$. $d_{\max}$ represents the maximum depth of the tree and is a free parameter: we set both $n$ and $d_{\max}$ to satisfy $n=2^{d_{\max}}$.

Let note $n_{dkt}$ the number of elements of $S_t$ belonging to the cell $Z_d^k$. By construction the root cell contains all the elements: $n_{00t}=n$. The isolation tree is built recursively according to the following rules: (i) if $n_{dkt}=1$, $Z_d^k$ is a leaf node and is not split further; (ii) if $d=d_{\max}$, $Z_d^k$ is also a leaf and is not split, even if possibly $n_{dkt}>1$; (iii) else a feature $f$ is randomly selected and a split point $x_{fdkt}^s$ is uniformly randomly drawn from the range $[\min \{ x_{fj} | x_j\in S_t \cap Z_d^k\} , \max \{ x_{fj} | x_j\in S_t \cap Z_d^k\} ]$. The cell $Z_d^k$ is then split into two daughter cells $Z_{d+1}^{2k} = Z_d^k \cap \{x_f\le x_{fdkt}^s\}$ and $Z_{d+1}^{2k+1} = Z_d^k \cap \{x_f > x_{fdkt}^s\}$.

The anomaly score is built out of the expected depth $E[d_{i}]_t$ reached by a data sample $i$ in each tree $t$.
 The depth $d$ within one tree is defined as the number of edges traversed between the root node and the leaf node that is, $d_{it} = d\,|\,x_i\in Z_{dt}^k$ where $Z_{dt}^k$ is a leaf. However, since the tree is cut at a maximum depth \( d_{\max} \), a leaf may contain more than one element of \( S_t \). In this case, we replace $d_{\max}$ with the expected depth that would be obtained if the tree were recursively built until each node contained a single element. The expected depth for $n$ sample is given by 
$E[d(n)] = 2H (n-1) - 2 (n-1)/n$, where $H(n)$ is the harmonic number. As a result, the total expected depth is $E[d_{i}]_t = d_{it} + E[d(n_{d_ik_{di}t | x_i\in Z_{dt}^{k}})]$.

The anomaly score for the isolation forest is then constructed from the average depth for each tree $<E[d_{i}]_t>$ and normalized by the expected depth E[d(n)]. With respect to the original definition, we add a $-$ sign for the anomaly score, without loss of generality: 
 $s(i)= -2^{-\frac{<E[d_i]_t>}{E[d(n)]}}$. Outliers are then defined as data points in underdense regions, characterized by a low anomaly score.

The score provided by isolation forest cannot be used directly to provide an explanation: anomalies are located in shallow branches of the tree, that is, underdense regions of the parameter space. However, it is possible by reworking the anomaly score to link it to the  features that contributed the most to it. We call this feature signature of the anomaly.

In order to build this signature, we rewrite the depth within the tree as follows:
\begin{eqnarray}
E[d_{i}]_t  = & d_{it} + E[d(n_{d_ik_{di}t})] \\ 
    = & (d-1)_{it} +E[d(n_{(d-1)_ik_{(d-1)i}t}] \nonumber \\
    & +1 - E[d(n_{(d-1)_ik_{(d-1)i}t}] + E[d(n_{d_ik_{di}t})] 
\end{eqnarray}
with $(d-1)_i$ and $k_{(d-1)i}$ are such as $x_i \in Z_{(d-1)t}^{k_{(d-1)}}$, that is, the mother cell from $Z_{dt}^{k}$.
In this formula, the first line of the second equation describes the expected depth after $d-1$ splits, while the next line corresponds to the effect of the $d$-th split. As this  split for event $i$ is linked to the selected feature $f_{(d-1)_i,k_{(d-1)i}t}$, we index it by the feature number $f$. Introducing the individual signature element as 
\begin{equation}
\delta S_{f_{(d-1)_i,k_{(d-1)i}t}} = +1 - E[d(n_{(d-1)_ik_{(d-1)_i}t}] + E[d(n_{d_ik_{di}t})] 
\end{equation}
the expected depth then can be rewritten as
\begin{equation}
E[d_{i}]_t  = E[d(n)] + \sum_d \delta S_{f_{(d-1)_i,k_{(d-1)i}t}}
\end{equation}

We then define the signature for feature $f^*$ and sample $i$ as 
\begin{equation}
S_{f^*i}=<\delta S_{f_{(d-1)_i,k_{(d-1)i}t}} > _{f=f^*}
\end{equation}
where the average is taken on the number of times the feature $f^*$ is selected when descending the trees for the sample $i$. Features responsible for separating outliers will have $n_{dk_dt} \ll  n_{(d-1)k_{d-1}t}$, which will result in a negative signature $S_f$, while positive signatures will pinpoint features that are ineffective in detecting $i$ as an outlier
.

The code implementing the feature signatures is available under the Coniferest package \cite{2024arXiv241017142K}.

\section{Signatures in action}

\begin{figure}[t]
\centerline{\includegraphics[width=0.9\linewidth]{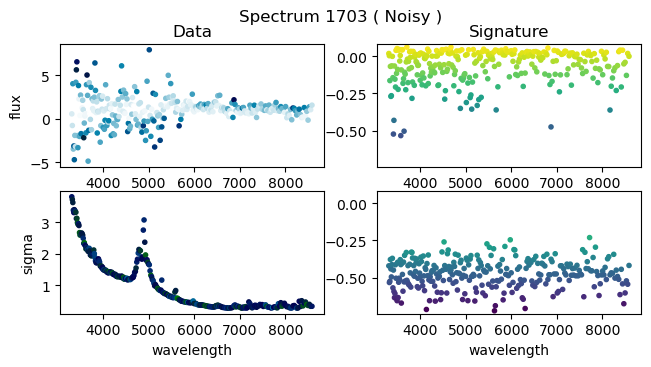}}
\caption{The most anomalous spectra according to isolation Forest. \textit{Left:} the original data in flux (top) and associated error (bottom). The darkest points correspond to the features having the strongest anomaly signature. \textit{Right:} associated signatures for flux and uncertainty.}
\label{fig:SnSignature}
\end{figure}

\begin{figure}[t]
\centerline{\includegraphics[width=0.9\linewidth]{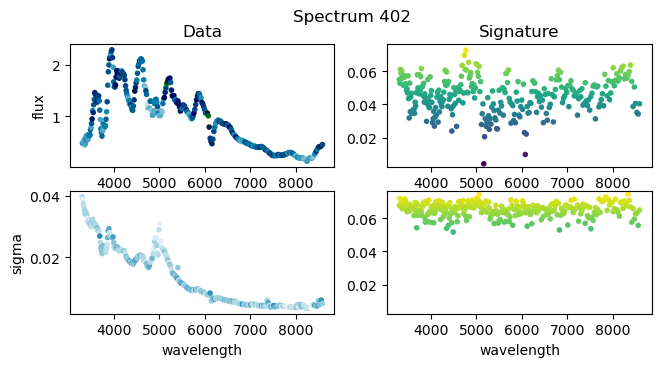}}
\caption{Same representation as figure 1, for the most nominal spectrum.}
\label{fig:SnSignatureNom} 
\end{figure}

Once the isolation forest is run on a data set, it is possible to compute the signatures for every sample. In this section we will illustrate two possible uses of signatures: visualization of the features contributing to the anomaly score, and finding new types of outliers. Finding similar outliers once one is selected is another possible use of signatures. 

\subsection{Outlier visualization}

We trained an isolation forest on the SNFactory data sample, with $n=1024$ and $T=3000$, and then computed the signatures for all spectra. Figures \ref{fig:SnSignature} and \ref{fig:SnSignatureNom} present the data and the signatures for the most outlier and the most nominal event in the sample. While human may spot immediately that the outlier has a high noise level, the signature shows that the decision to label it as an outlier is made on the variance spectrum, rather than on the noisy data themselves. For the nominal spectrum, one may see that the signatures are positive, indicating that no feature is isolating this event as anomalous.

\begin{figure}[t]
\centerline{\includegraphics[width=0.6\linewidth]{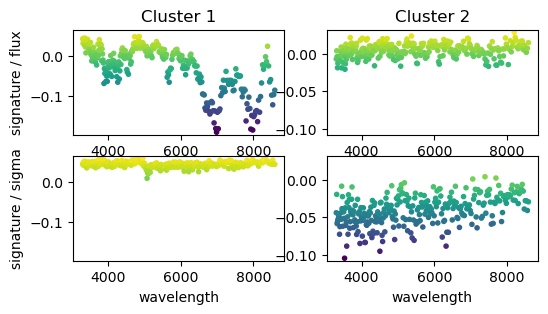}}
\includegraphics[width=0.9\linewidth]{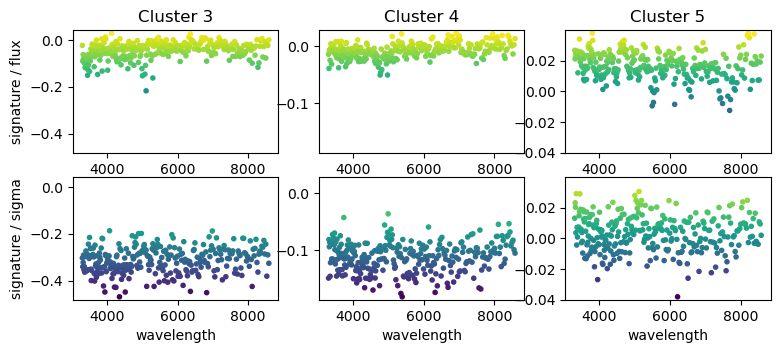}
\caption{Average signatures for each K-means cluster performed on the top 10\% anomalies. For each cluster the signature relative to the flux \textit{(top)} and uncertinaty \textit{(bottom)} is presented. Cluster 1 differ signigicantly from others and therefore is suspected to contain different classes of anomalies.}
\label{fig:SnKmeans}
\end{figure}

\subsection{Searching for different outliers}

A typical case
when dealing with noisy data is when the outlier detection algorithm keeps finding noisy events, while missing the interesting ones.

With signatures, it is possible to speed-up the search for different anomalies. Here, we performed a K-means with $K=5$ on the top 10\% most anomalous objects detected in the SNFactory sample. Figure \ref{fig:SnKmeans} shows that the average signature is significantly different from the noise signature for only 1 out of 5 clusters. The expert then only has to scan 39 spectra out of 232 to find new anomalies, speeding-up anomaly discovery rate by a factor 6 in this example.

\section{Conclusion}
We presented a new metric for feature importance in the context of outlier detection: the Signature. This metric enables new tasks when searching for outliers: visualization of the isolation forest decision, finding different kind of anomalies, or finding more outliers sharing a similar signature pattern. The use of Signatures is not restricted to astronomical use cases, and can be employed for all tabular data once an isolation forest is applied.

\begin{acknowledgement}
Support was provided by Schmidt Sciences, LLC. for K. Malanchev. M.~Kornilov, A.~Lavrukhina, M.~Pruzhinskaya, T.~Semenikhin acknowledge support from a Russian Science Foundation grant 24-22-00233, https:
//rscf.ru/en/project/24-22-00233/. 
\end{acknowledgement}

\bibliographystyle{spmpsci}
\bibliography{biblio}

\end{document}